\documentclass[10pt,twocolumn,letterpaper]{article}

\usepackage{iccv}
\usepackage{times}
\usepackage{epsfig}
\usepackage{graphicx}
\usepackage{amsmath}
\usepackage{amssymb}
\usepackage{subcaption} 
\usepackage{enumitem}
\usepackage{array}
\usepackage{adjustbox}

\usepackage[breaklinks=true,bookmarks=false,hidelinks]{hyperref}

\usepackage[capitalize]{cleveref}
\crefname{section}{Sec.}{Secs.}
\Crefname{section}{Section}{Sections}
\Crefname{table}{Table}{Tables}
\crefname{table}{Tab.}{Tabs.}
\usepackage{lipsum}

\newcommand\blfootnote[1]{%
  \begingroup
  \renewcommand\thefootnote{}\footnote{#1}%
  \addtocounter{footnote}{-1}%
  \endgroup
}

\iccvfinalcopy 

\def\iccvPaperID{****} 

\ificcvfinal\fi

\begin{document}

\title{OSPC: Online Sequential Photometric Calibration}

\author{Jawad Haidar\\
{\tt\small jsh19@mail.aub.edu}
\and
Douaa Khalil\\
{\tt\small dk69@aub.edu.lb}
\and
Daniel Asmar\\
{\tt\small da20@aub.edu.lb}
}

\maketitle
\ificcvfinal\thispagestyle{empty}\fi

\begin{abstract}
Photometric calibration is essential to many computer vision applications. One of its key benefits is enhancing the performance of Visual SLAM, especially when it depends on a direct method for tracking, such as the standard KLT algorithm. Another advantage could be in retrieving the sensor irradiance values from measured intensities, as a pre-processing step for some vision algorithms, such as shape-from-shading. Current photometric calibration systems rely on a joint optimization problem and encounter an ambiguity in the estimates, which can only be resolved using ground truth information. We propose a novel method that solves for photometric parameters using a sequential estimation approach. Our proposed method achieves high accuracy in estimating all parameters; furthermore, the formulations are linear and convex, which makes the solution fast and suitable for online applications. Experiments on a Visual Odometry system validate the proposed method and demonstrate its advantages.
\end{abstract}
\blfootnote{The paper is under consideration at Pattern Recognition Letters}
\section{Introduction}

\label{sec:intro}
Photometric calibration consists of estimating the mapping between the scene radiance and the camera intensity values \cite{grossberg2003space}. The main factors that affect this mapping, in addition to the scene radiance, are the camera response function, its exposure values, and the vignette. The mapping is nonlinear due to the camera response function (CRF) that translates irradiance to camera intensity values. Auto exposure cameras help in capturing regions with high (low) brightness by lowering (increasing) the value for exposure. Although auto-exposure does enhance the dynamic range of a camera, it also complicates the mapping estimation. For example, if we consider two images of the same static object, taken by a stationary camera at times $t_1$ and $t_2$, then the two images will have the same intensities when the scene radiance at $t_1$ is double that at $t_2$, and the first exposure value is half the second. 

\begin{figure}
     \centering
     \begin{subfigure}[b]{0.22\textwidth}
         \centering
         \includegraphics[width=\textwidth]{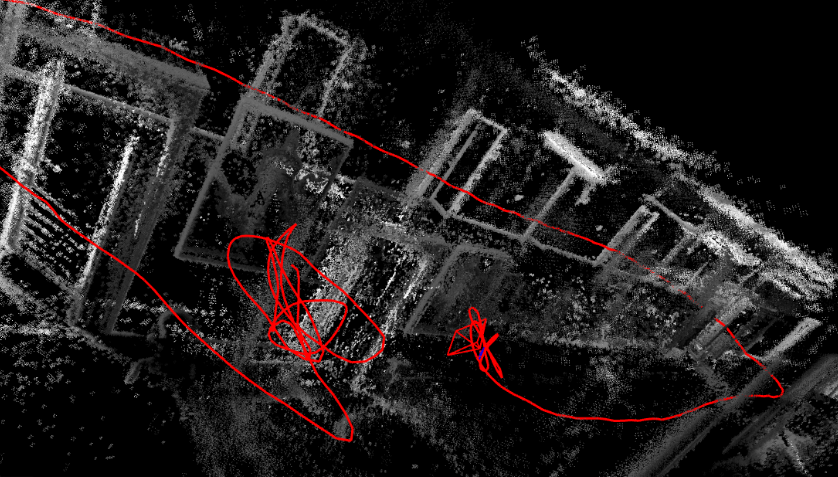}
         \caption{Without calibration.}
         \label{subfig:wo_calib_seq36}
     \end{subfigure}
     \hfill
     \begin{subfigure}[b]{0.22\textwidth}
         \centering
         \includegraphics[width=\textwidth]{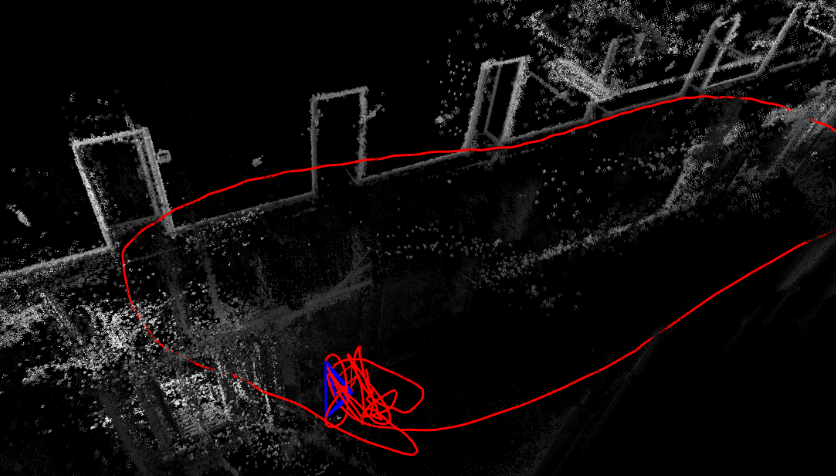}
         \caption{With our calibration.}
         \label{subfig:w_our_calib_seq36}
     \end{subfigure}
        \caption{Qualitative result showing the reconstructed map of TumMono (Sequence 36) without and with our photometric calibration. A drastic drift is shown in the absence of calibration, notably between the start and end segments, compared to a smaller one when the sequence is photometrically calibrated.}
        \label{fig:qualitative-result}
\end{figure}
Additionally, the vignette causes irradiance attenuation that increases radially, thereby adding a spatial dimension to the mapping. Hence, we would have different intensity values for the same object under the same radiance and exposure when the camera moves.  

Photometric calibration has several benefits. For one, it can help relax the constant brightness assumption that is the foundation of widely used feature trackers like Kanade-Lucas-Tomasi (KLT) and other direct algorithms \cite{kanadeiterative,323794}. Furthermore, photometric calibration is essential for algorithms that require knowledge of the scene radiance, such as shape from shading, photometric stereo, high dynamic range image formation, and color constancy \cite{1541349}. 

Most of the existing photometric calibration systems that solve for all the parameters use a joint approach. However, in addition to being a computationally expensive method, it encompasses an ambiguity that is often addressed by using ground truth information. The goal of this paper is to perform an online estimation of the photometric parameters. We tackle this problem as a sequential optimization problem that estimates the camera response function, vignette, and exposure values independently. This approach results in avoiding the ambiguity problem.

The remainder of this paper is organized as follows. In the next section, preliminary concepts of photometric calibration are explained, followed by a review of the previous methods and their shortcomings. In \cref{sec:Method}, a novel method for estimating the camera response function, vignette, and exposures is presented. We evaluate our approach in \cref{sec:exp} with various experiments, followed by an ablation study in \cref{sec:ablation}, and finally conclude in \cref{sec:conc}. 

\section{Background and related work}
\label{sec:relatedwork}
When the light from a source is reflected on a surface, part of that reflected light penetrates the lens of the observing camera. Radiance is a measure of the rate at which light energy is emitted from a surface in a particular direction, it is often denoted by $L$. If the scene surface follows Lambert's Law, then the emitted radiance is independent of the direction it was emitted from, and it is assumed to be the same with respect to the observer's viewing angle.  

The total light received by the sensor of the camera, as a function of sensor spatial location $x \in \Omega$, is called irradiance $I(x)$. Most of the time, when rendering an image of an object or scene, the amount of light incident at each sensor element, or in other words, the irradiance, is attenuated by a vignetting factor $I(x) = V(x)L$, where $V: \Omega \rightarrow [0, 1]$, which can be caused by optical and mechanical effects. Eventually, in real imaging systems, the vignette causes the image brightness to be reduced at the periphery compared to the center of the image, implying that scene points with the same emitted radiance but in different directions, won't possess the same captured irradiance by the sensor.

Furthermore, the accumulated irradiance at a sensor element location depends on the amount of time, called exposure time $e$, for which the camera film or sensor was actually exposed to the incident light. Assuming that $I(x)$ is constant over this time window, we can write $I_{acc}(x) = eI(x)$. The CRF, $f : \mathbf{R} \rightarrow [0, 255]$, then maps $I_{acc}$  to an output intensity value. In most cameras, this function is non-linear, and it's intentionally designed this way by camera manufacturers for different purposes \cite{4587648}, such as compressing the dynamic range of scene radiance so the output can be fixed to camera range $[0, 255]$ or normalized intensity $[0, 1]$. Finally, a scene point radiance $L$ can be mapped to the measured intensity value $M$ by the equation: $M = f(eV(x)L)$.

To date, a number of photometric calibration methods have been proposed. Lin \etal \cite{lin2004radiometric} introduced a method to recover the camera response function, based on the relationship between the inverse response and the non-linear distribution of measured colors around edge regions in a single RGB image. However, this color-based approach is not applicable to grayscale images that are commonly used in vision applications. In \cite{lin2005determining}, Lin and Zhang recovered the camera response from a single grayscale image, using the skewed histogram distribution of the measured pixel intensities around edge regions. Their method takes advantage of the direct relationship between this skewness and the non-linearity of the camera response function. For better accuracy, both of these methods employed the empirical model of response (EMoR) to estimate the inverse response. The EMoR was compiled by Grossberg and Nayar from the real-world camera response database \cite{grossberg2003space}. Although their approaches rely on single images, allowing them to process pictures taken by unknown or unavailable cameras, they do not recover the vignette nor the exposure times. Likewise, the authors in \cite{mo2019ambiguity} based their method on the non-linear distribution of color edges in order to estimate ambiguity-free CRFs of various online photos. However, they do so while ignoring the effect of vignette. An attempt to recover the vignetting function from a single image was established by Zheng \etal \cite{zheng2008single}, identifying, through iterative segmentation techniques, pixels having the same scene radiance, then measuring their photometric falloff towards the image borders. This method expects a CRF to be calculated as a pre-processing step, and just like \cite{lin2004radiometric,lin2005determining}, it is only meant for vignette estimation and do not recover the exposure times. 

Recovering the camera response function or its inverse can also be achieved by acquiring multiple images of a static scene, taken under different exposures \cite{debevec2008recovering,mitsunaga1999radiometric}, using the relationship $g(M_A) = k * g(M_B)$, where $g$ is the inverse response, $k$ is the ratio of exposures between each two images, $M_A$ and $M_B$ are the measured intensities of two arbitrary corresponding points. Assuming known exposure times, the algorithm described in \cite{debevec2008recovering} recovers the inverse response using a non-parametric model, then uses it to construct a map of the radiance in the scene, up to a scale. Authors of \cite{mitsunaga1999radiometric} used a high-order polynomial model to iteratively solve for both the inverse response function and the actual exposure ratios, starting from rough estimates of the ratios in addition to pixel intensity measurements. After finding the camera response function based on \cite{debevec2008recovering}, Engel \etal \cite{engel2016photometrically} estimate the vignette using a non-parametric model from a sequence of images showing a planar, Lambertian, and non-uniformly colored scene. All these static-based methods to recover the response function do not apply to online and offline dynamic videos, in which objects and/or cameras are in motion. Therefore, it is necessary to use an algorithm that can photometrically calibrate arbitrary video sequences.

Kim \etal \cite{kim2010joint} employ a multi-scale iterative method to jointly estimate the camera response function, exposure ratio, and pixel displacements. Due to their small displacement assumption between consecutive frames in videos, their algorithm ignores the effect of vignette. In addition, in order to deal with the ambiguity problem that arises when solving jointly for photometric parameters \cite{1240119}, they chose to set the value of the response function at a certain image intensity value, extracted from ground truth data which is not available in real-world scenarios. A more recent work \cite{bergmann2017online} jointly recovers the exposure times of consecutive frames, the camera response function, and the vignetting function for auto-exposure videos, in both online and offline fashions. However, just like in \cite{kim2010joint}, they use a point from the camera response function ground truth data in order to solve the ambiguity. 

Authors of \cite{kim2008robust} proposed a semi-sequential approach to photometric calibration, by decoupling the vignette estimation from both the camera response function and exposure difference estimation. Their algorithm starts by selecting corresponding points with specific properties depending on which estimation process they are running, but they fall back again into the exponential ambiguity problem when solving jointly for both the camera response function and the exposure difference. Instead of relying on response function ground truth data to solve the ambiguity, they chose to make an assumption on the value of the exposure difference between the first image pair, by setting it to a constant, arguing that the choice of this constant is not critical for the applications they are targeting. Furthermore, their approach relies on aligning image pairs, which allows a large number of corresponding points but slows down the run-time of the application, making it unsuitable for online calibration.

In our approach, we adopt a fully-sequential online photometric calibration method. We solve for each photometric parameter independently, taking advantage of the exposure metadata retrieved along with the images in order to make our solution ambiguity-free without targeting specific applications such as in \cite{kim2008robust}. 
The main contributions of this paper are summarized as follows:
\begin{itemize}
  \item A novel approach for photometric calibration that overcomes the exponential ambiguity that previous approaches faced by utilizing exposure metadata obtained from cameras in real-world settings, eliminating the need for CRF ground truth data.
  \item Formulating all the estimation problems in a convex optimization framework, yielding a quick and computationally efficient estimation of the parameters. 
  \item Experimental results demonstrate accuracies that are at par with (or better than) state-of-the-art in estimating photometric parameters.
\end{itemize} 

\section{Methodology}
\label{sec:Method}
\begin{figure}[ht!]
    \centering
    \includegraphics[width=\linewidth]{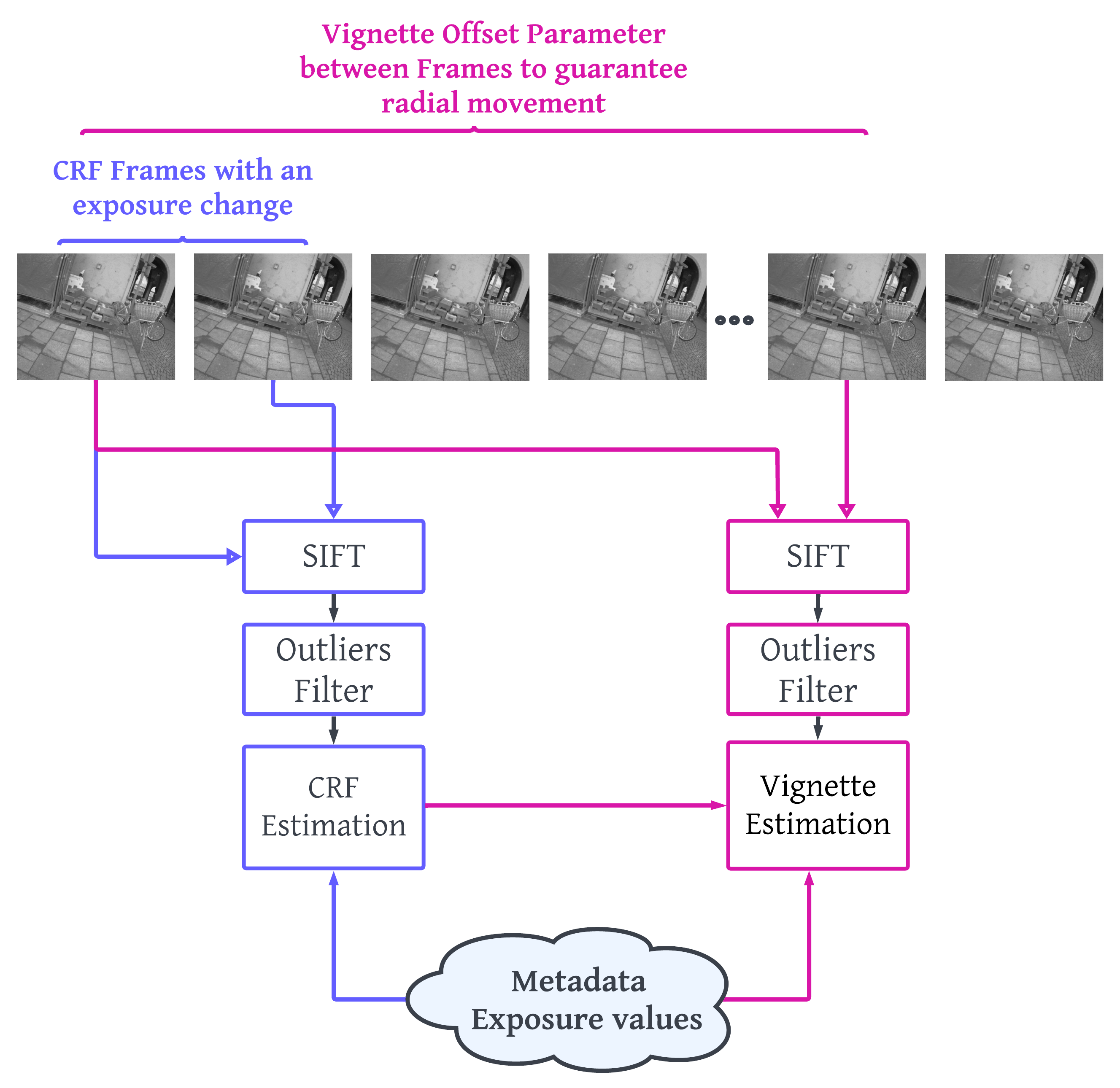}
    \caption{Overview of the main steps of our OSPC approach.}
    \label{fig:flowchart}
\end{figure}

The main inputs for a photometric calibration objective function are the corresponding points in frames. Earlier approaches have used direct methods to obtain the point correspondences. In \cite{grundmann2013post}, the authors used the pyramidal implementation of KLT; however, this implementation is not robust to drastic exposure changes between consecutive frames. The authors in \cite{bergmann2017online,kim2010joint}, overcame this issue by using gain adaptive KLT that jointly estimates the displacements, camera response function, and the exposure ratio. Although this approach provides good corresponding points, the joint estimation leads to exponential ambiguity, which causes infinite solutions for the same objective function; in other words, each estimated parameter will be raised to an unknown exponential value (\eg $\tilde{f}(x)=f(x^{\frac{1}{\gamma}})$, $\tilde{V}(R)=V(R)^{ \gamma} $, and $\tilde{e}=e^{\gamma}$). This additional gamma parameter can only be determined once prior knowledge on the camera response function is provided.
Our approach is to formulate the photometric calibration as a sequential estimation problem, given that the authors in \cite{1240119} mathematically proved that sequential photometric calibration is free of exponential ambiguity. \cref{fig:flowchart} presents the different steps of our current implementation.

\subsection{Tracking}
\label{subsec:Tracking}
Standard direct tracking methods are not robust to photometric changes due to the constant intensity assumption. Consequently, we perform tracking using Scale Invariant Feature Transform (SIFT) \cite{lowe2004distinctive} to improve robustness. Additionally, each estimation formulation requires unique keyframes; for instance, the keyframes used for estimating the camera response function are required to have a very small radial displacement of features, while requiring a large change in exposure. On the other hand, the vignette estimation requires keyframes with high radial movement in order to capture spatial attenuation.

To remove potential outliers from the corresponding pairs, we rely on a filter that is based on the assumption that features exhibit the same motion locally. The image is divided into equally sized blocks, in each block the distribution of the orientations of displacement vectors is binned. The corresponding points that belong to the bin with the highest frequency are used as inputs to the optimization problem.
  
\subsection{Optimization}
\label{subsec:Optimization}
In our approach, we leverage the metadata to first extract the exposure values. This is an essential pillar that helps in transforming the estimation from a joint into a sequential prediction.\\
\textbf{Camera response function.} The estimation of the CRF starts by searching for pairs of frames with an exposure change that is high enough to guarantee the invertibility of the matrix in \cref{objective}. For instance, if the exposure ratio is equal to one, the elements in the matrix would all be equal or close to zero. As the exposure ratio changes, it is more probable for the system to have independent vectors or, in other words, a full rank. The radial distance is then calculated for all corresponding pairs and only those with small radial movement are chosen. These corresponding points have in common the same scene radiance, assuming they lie on a Lambertian surface. Consequently, by dividing the irradiance of each pair, the scene radiance cancels out, as well as the vignette. The cancellation of the vignette is valid because small radial displacement causes negligible changes in the vignette value. The irradiance division is shown in \cref{I_div}. 
\begin{equation} 
\label{I_div}
\frac{f^{-1} (M_1)}{f^{-1} (M_2)}  = \frac{e_1}{e_2},
\end{equation} 
where \(f^{-1}\) stands for the inverse camera response function, $M1$ and $M2$ are the corresponding intensity values, and $e$ is the exposure value. The inverse CRF is modeled as a polynomial of degree 2, which is a good approximation in terms of accuracy
and the time needed to find the three parameters \cite{grossberg2003space}. Subbing \((c_0 + c_1M + c_2M^2)\) into \cref{I_div} we have:
\begin{equation} 
\label{poly_div}
\frac{c_0 + c_1M_1 + c_2M_1^2}{c_0 + c_1M_2 + c_2M_2^2}  = \frac{e_1}{e_2}
\end{equation} 
Reformulating \cref{poly_div} in the vector form:
\begin{equation}
    \label{matrix_form}
    \begin{bmatrix} 
      1-k & M_1-kM_2 &  M_1^2-kM_2^2    
    \end{bmatrix} 
    \begin{bmatrix}
     c_0 \\
     c_1 \\
     c_2 \\
    \end{bmatrix}
     = 0,
\end{equation}
where $k$ is the exposure ratio. Each pair of corresponding points adds a new row to the system, which eventually leads to a set of homogeneous equations with an infinite number of solutions. However, all the camera response functions belong to the space of functions that undergoes the constraints $ f^{-1}(0)=0 $ and  $ f^{-1}(1)=1 $. Consequently, adding these equality constraints leads to a unique solution, where the objective function can be formulated as follows:
\begin{align}
\label{objective}
\min_{c} \quad & \lVert Ac \rVert^2 \\
\textrm{s.t.} \quad & \sum_{i=1}^{i=N}{c_i}=1 \\
&c_0=0,
\end{align}
where \textit{A} stands for the left matrix in \cref{matrix_form} for all the samples, and \textit{c} is a vector of the CRF parameters. This can be further written in a more compact form:
\begin{align}
\label{objective_compact}
\min_{c} \quad & c^TA^TAc \\
\textrm{s.t.} \quad & Ec=d,
\end{align}
\textit{E} consists of two rows, with the first row being full of ones, and the second row having only the first element as 1 while the remaining are equal to zero. Additionally, \textit{d} consists of two elements: the first is 1 and the second is 0.

To solve this optimization problem, we minimize the Lagrangian function:
\begin{align}
\label{equ:lagrangian}
\min_{c,\lambda} \quad & c^TA^TAc + \lambda^T(Ec-d)
\end{align}

Calculating the second derivative with respect to \textit{c} leads to a positive semi-definite matrix $A^TA$. This confirms that this formulation is convex; therefore, there exists a global minimum to the problem that can be reached by setting the gradients, with respect to \textit{c} and \textit{$\lambda$}, to zeros.  The final form can be written as:
\begin{equation}
    \label{final_form}
    \begin{bmatrix} 
      2(A^TA)  & E^T \\
      E        & 0        
    \end{bmatrix} 
    \begin{bmatrix}
     c \\
     \lambda \\
    \end{bmatrix}
     = \begin{bmatrix}
      0 \\
      d
     \end{bmatrix}
     ,
\end{equation}
which can be easily solved using a pseudo inverse.\\
\textbf{Vignette.} The frames used to extract the corresponding points for the vignette estimation should include high radial motion. To guarantee this constraint, we extract corresponding pairs from frames that are not consecutive. In other words, we select Frame One as it arrives, then skip a number of frames before choosing the second one. 

Similar to the formulation of the camera response function, the estimation of the vignette parameters starts by dividing the irradiance of corresponding points. However, as seen in \cref{V}, the vignette is included in the equation due to the presence of a high radial movement.
\begin{equation}
    \label{V}
    \frac{f^{-1}(M_1)}{f^{-1}(M_2)}=\frac{e1}{e2}\frac{V(R_1)}{V(R_2)}
\end{equation}

To this end, the only remaining unknowns are the parameters of the vignette model, as we will use the estimated camera response function and extract the exposure value from the image metadata. The radial model is often employed in the literature for vignette correction, assuming that the vignette attenuation is symmetrical and centered at the frame's center.  
\begin{equation}
    \label{radial}
    \frac{ 1+R_1^2v_1 + R_1^4v_2 + R_1^6v_3    }{1+R_2^2v_1 + R_2^4v_2 + R_2^6v_3}=\frac{f^{-1}(M_1)}{f^{-1}(M_2)}\frac{1}{k}
\end{equation}
After performing simple algebraic manipulation, the estimation problem can be written in matrix form shown in \cref{martix_v}.
\begin{equation}
    \label{martix_v}
    \begin{bmatrix}
     R_1^2-\psi R_2^2 & R_1^4-\psi R_2^4 & R_1^6-\psi R_2^6
    \end{bmatrix}
    \begin{bmatrix}
     v1 \\
     v2 \\ 
     v3
    \end{bmatrix}
    = \psi - 1
\end{equation}
\textit{k} and $\psi$ stand for exposure ratio and $\frac{f^{-1}(M_1)}{f^{-1}(M_2)}\frac{1}{k}$ respectively. Solving for the vignette parameters is fast since the formulation is linear and can be solved using least squares.\\
\textbf{Exposure.} Once we get a CRF and a vignette estimation, we can judge their accuracy using re-computed exposures with this formulation: 
\begin{equation}
    \label{equ:mean}
    k=\frac{1}{N}\sum_1^N\frac{f^{-1}(M_1)V(R_2)}{f^{-1}(M_2)V(R_1)},
\end{equation}
in which \textit{N} stands for the total number of corresponding points per image pair. The mean in this estimation is necessary to reduce the effect of white noise. The estimated exposure ratio is to be compared with the available metadata at each keyframe pair used in the estimations. The validation of the CRF and the vignette will be performed sequentially based on how close the estimated exposure is to the metadata. When the CRF accuracy is considered, the vignette ratios would be ignored as in \cref{I_div}.
 
\section{Experiments and analysis}
\label{sec:exp}
To evaluate our model, we used some sequences from the TumMono dataset \cite{engel2016photometrically}, which includes videos recording various environments both indoor and outdoor. Additionally, the ground truth photometric parameters are all provided with each sequence, which allow us to first directly compare our estimates against the truth and another calibration method, and second, to compare the performance of a Visual Odometry system in terms of trajectory error when given ground truth versus estimated photometric parameters as inputs. Both of these objectives help us judge the accuracy of our estimation.
Unless stated otherwise, the denotation of CRF in this section refers to the inverse camera response function $f^{-1}$ instead of \textit{f}. 

\subsection{ Photometric estimates }
As mentioned previously, we start by estimating the parameters of the CRF. To this end, only a few pair of keyframes are needed; however, the constraints on choosing these pairs (\ie, the high exposure change and small radial displacement) are not always found across the entire sequence. Consequently, to guarantee a sufficient number of keyframe pairs for the estimation problem, we select those that verify the constraints from a window of 200 frames.

The algorithm starts by scanning the meta exposure values of these 200 consecutive frames as they arrive: when the exposure ratio for an image pair is less than \textit{0.92} or more than \textit{1.08}, the image pair is considered for further processing, thus adding two keyframes to the system; otherwise, it is ignored. SIFT features are then extracted from each keyframe. We proceed with outlier removal among the found matches, using a filter described in \cref{subsec:Tracking}. Finally, only the corresponding points that feature a small radial displacement are selected, which can be found in local parts where the scene did not change much. At this point, all the required data to solve the linear constrained optimization problem is available. 

Having the estimated CRF with the meta-exposure values as inputs, our method can now solve for the vignette parameters. As described in \cref{subsec:Tracking}, this estimation requires a large radial displacement of features, which can be potentially found in a pair of distant frames, rather than consecutive ones. In our experiments on some TumMono sequences, we found that an interval of 30 frames is sufficient to guarantee radial motion. Hence, after selecting and saving the first frame, we save the next 29 frames in a database as they arrive. Then, features are extracted and matched between the first and the $30^{th}$ frames. If the pair of images feature a large radial movement, a new row is added to \cref{martix_v}; otherwise, the pair does not contribute to the estimation. In both cases, the first frame in the database (\ie, the oldest) gets discarded, and the currently retrieved one from the camera is pushed back into the database. The new first and $30^{th}$ frames are now examined. This process is repeated over a window of total 400 frames, by offsetting the 30 frames by one at frame rate, accumulating sufficient data for the vignette estimation. The overall vignette \cref{martix_v} is represented as a least squares problem, yielding a fast estimation.

We repeat the process of estimating each of the CRF and vignette several times throughout the video sequence, then we average the results.

\begin{figure*}
     \centering
     \begin{subfigure}[b]{0.4\textwidth}
         \centering
         \includegraphics[width=\textwidth]{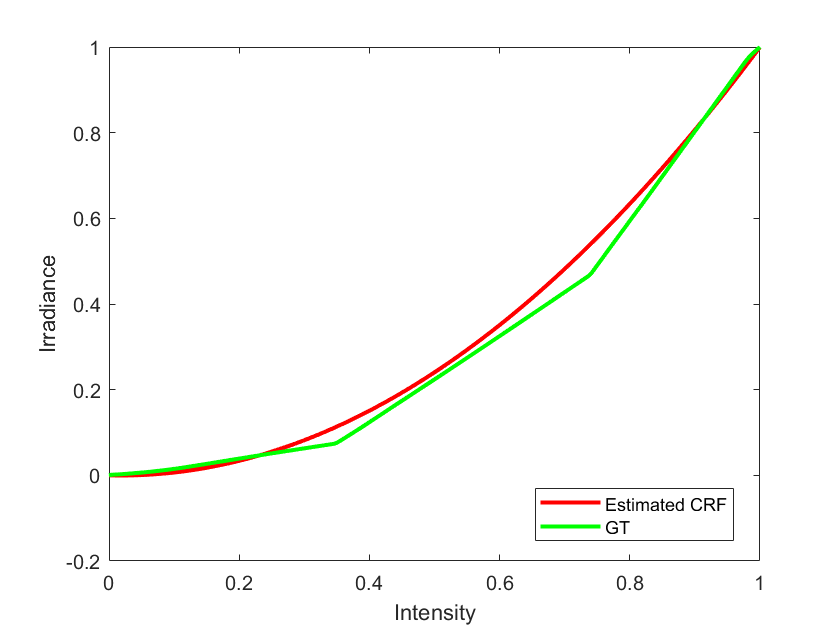}
         \caption{CRF}
         \label{subfig:our_crf_est}
     \end{subfigure}
     \begin{subfigure}[b]{0.4\textwidth}
         \centering
         \includegraphics[width=\textwidth]{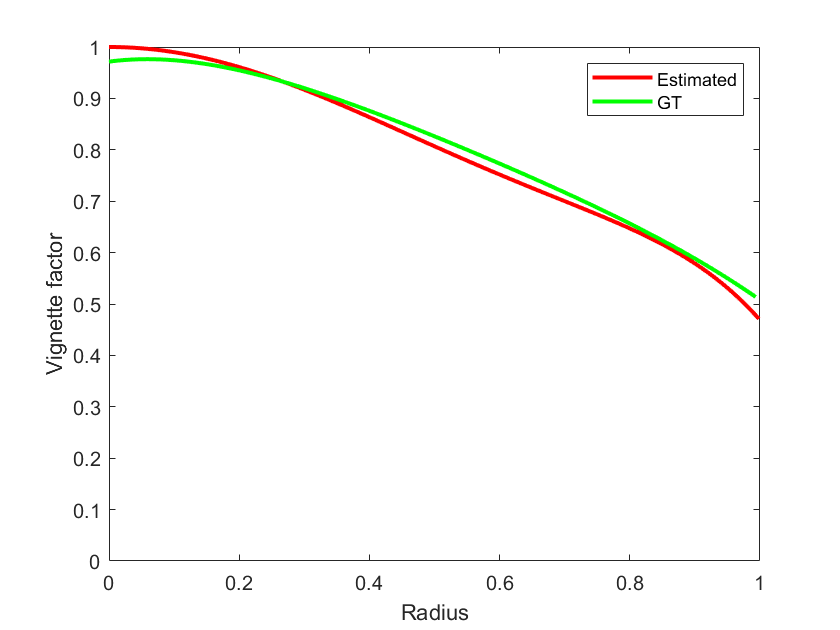}
         \caption{Vignette}
         \label{subfig:our_v_est}
     \end{subfigure}
          \hfill
     \begin{subfigure}[b]{0.5\textwidth}
         \centering
         \includegraphics[width=\textwidth]{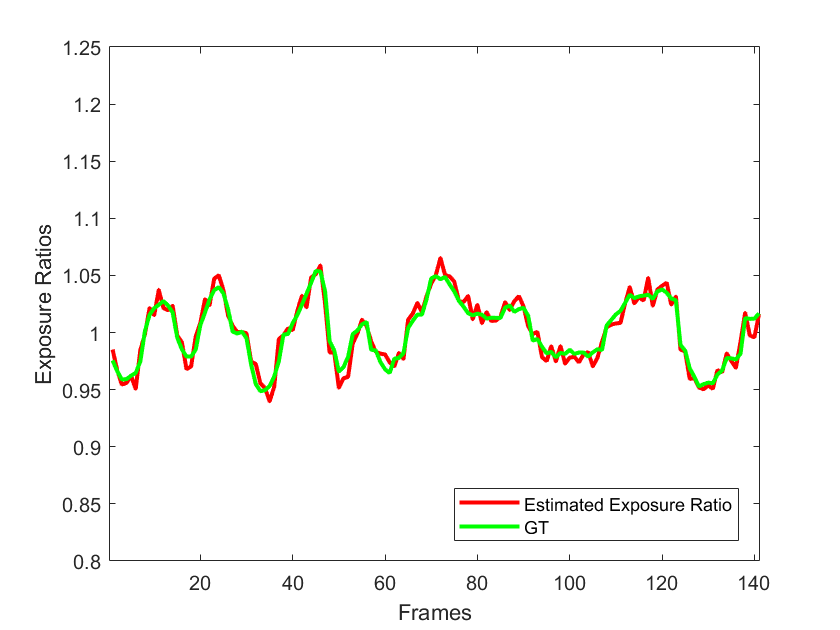}
         \caption{Exposure ratios}
         \label{subfig:our_exp_est}
     \end{subfigure}
        \caption{Our estimated photometric parameters.}
        \label{fig:our_ests}
\end{figure*}

\cref{fig:our_ests} shows the results we obtained after calibrating on Sequence 30. Our estimates are very close to the ground truth parameters, which proves the validity of the sequential formulation of the problem. In addition, the estimated exposure ratios being close to the metadata proves the efficiency of the validation method computed via \cref{equ:mean}. 

\begin{figure}
     \centering
     \begin{subfigure}[b]{0.34\textwidth}
         \centering
         \includegraphics[width=\textwidth]{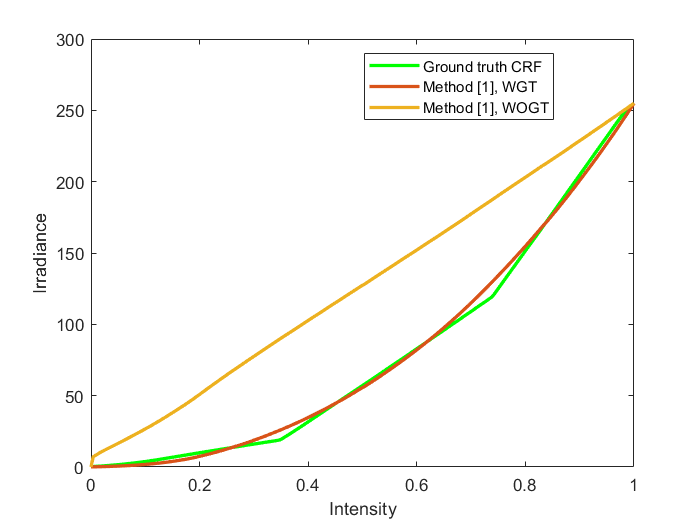}
         \caption{CRF online estimates.}
         \label{subfig:online_2_crfs}
     \end{subfigure}
     \hfill
     \begin{subfigure}[b]{0.34\textwidth}
         \centering
         \includegraphics[width=\textwidth]{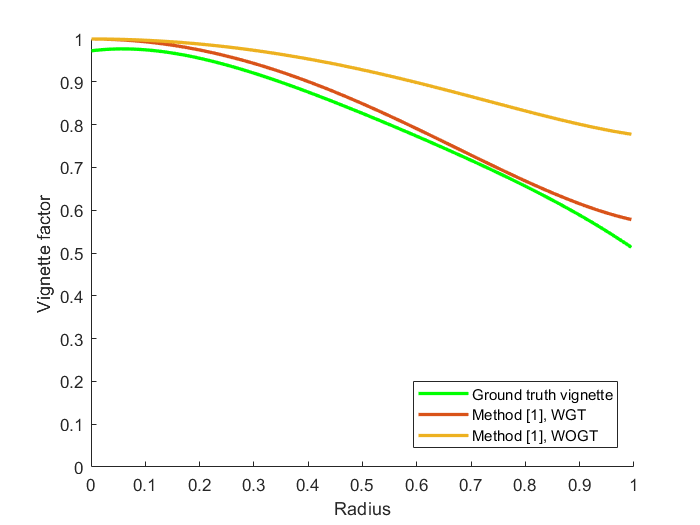}
         \caption{Vignette online estimates.}
         \label{subfig:online_2_vignettes}
     \end{subfigure}
        \caption{The online calibration method \cite{bergmann2017online} was evaluated by comparing the photometric estimates obtained using the CRF ground truth (WGT) and those obtained without it (WOGT).}
        \label{fig:their_est}
\end{figure}

The accuracy of the online joint estimation \cite{bergmann2017online} is reliant on one point from the ground truth; without this information, their estimates drastically diverge, as seen in \cref{fig:their_est}, due to the exponential ambiguity that arises once the photometric calibration is formulated as a joint optimization problem. This is our main advantage over previous approaches, as our system can be deployed in real scenarios where the ground truth data is not available, without significant degradation in accuracy, if any.     

\subsection{Experiment with visual odometry system}
To prove the importance of photometric calibration for visual Odometry (VO) or SLAM, we used the Direct Sparse Odometry system (DSO) \cite{engel2017direct} and the following two metrics: trajectory alignment error and reconstructed map quality. \cref{fig:qualitative-result} shows the map of TumMono Sequence 36 reconstructed by DSO with and without our calibration. To further show the applicability of our method to DSO and compare its accuracy against \cite{bergmann2017online}, we first ran the VO system with full ground truth photometric parameters, followed by these experiments, all on 24 TumMono sequences taken with the non-fisheye camera:
\begin{enumerate}[label=(\Roman*)]
\item \label{itm:expa} We ran DSO using our estimated photometric parameters as inputs.
\item \label{itm:expb} Using the open-source implementation of \cite{bergmann2017online}, we got their estimated photometric parameters in order to use them as inputs for the DSO system. 
\item \label{itm:expc} We disabled all the components of photometric calibration in DSO, including the affine brightness correction, to force a brightness constancy assumption as described in their parameter studies \cite{engel2017direct}.
\end{enumerate}

\begin{figure*}
     \centering
     \begin{subfigure}[b]{0.3\textwidth}
         \centering
         \includegraphics[width=\textwidth]{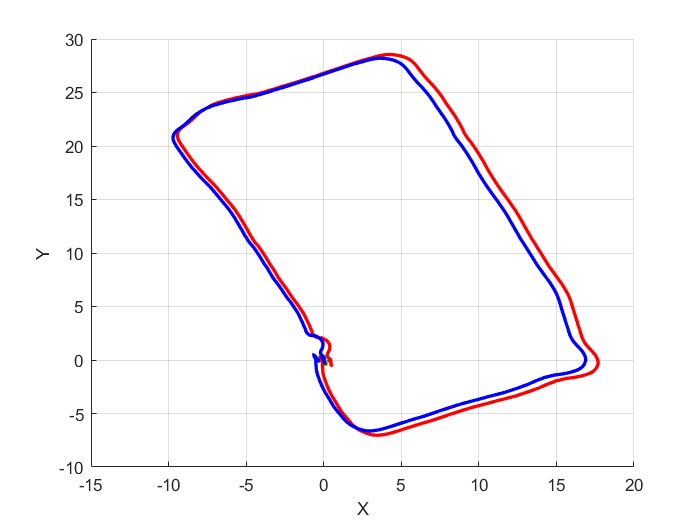}
         \caption{Experiment \ref{itm:expa}.}
         \label{subfig:seq50_ours}
     \end{subfigure}
     \begin{subfigure}[b]{0.3\textwidth}
         \centering
         \includegraphics[width=\textwidth]{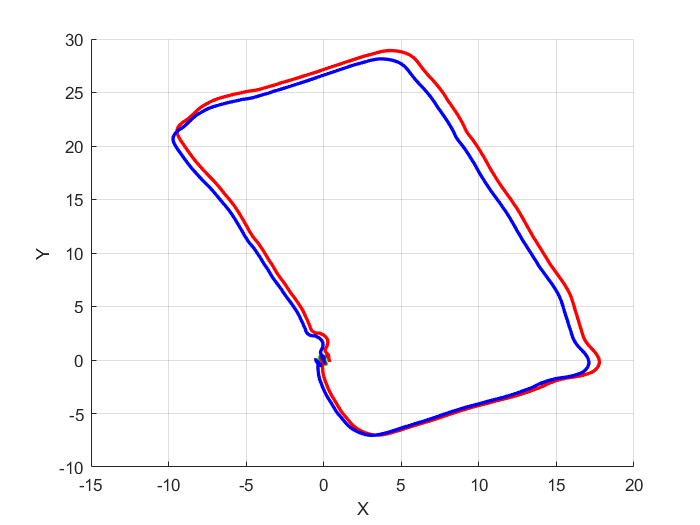}
         \caption{Experiment \ref{itm:expb}.}
         \label{subfig:seq50_online_wgt}
     \end{subfigure}
     \begin{subfigure}[b]{0.39\textwidth}
         \centering
         \includegraphics[width=\textwidth]{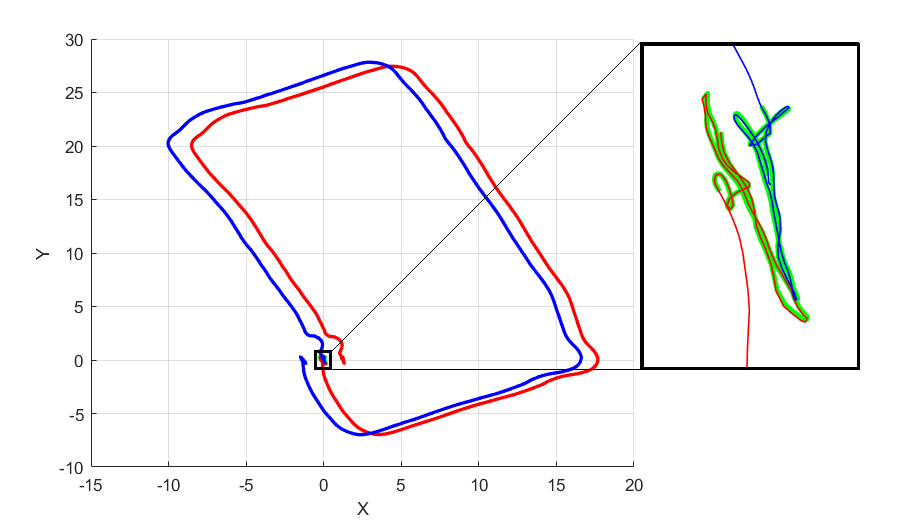}
         \caption{Experiment \ref{itm:expb} modified.}
         \label{subfig:seq50_online_wogt_with_closeup}
     \end{subfigure}
        \caption{Trajectories obtained from Experiment \ref{itm:expa}, Experiment \ref{itm:expb}, and Experiment \ref{itm:expb} modified. The small green-dashed area (around (0, 0)) are the set of available ground truth poses at the start and end segments. The tracked trajectory is as follows: (i) aligned to the start-segment in blue and (ii) aligned to the end segment in red. Ideally, without drifting, the aligned trajectories (i) and (ii) should overlap. The alignment error $e_{align}$ computes the RMSE between the red and the blue line, over the full trajectory. Notice that there is more drifting in (c) than in (a) and (b), which results in a higher $e_{align}$.}
        \label{fig:trajectories_expb_expc}
\end{figure*}

In the experiments above, we measured the trajectory alignment error ($e_{align}$ in meters) found in \cite{engel2016photometrically}. We performed 10 runs for each experiment to account for the probabilistic aspect of DSO, and we calculated their median. Since both of our and the baseline calibration \cite{bergmann2017online} methods make use of exposure metadata values, they were used in DSO instead of the estimated ones in all cases except Experiment \ref{itm:expc}, in which exposures are set to ones. Results are reported in \cref{table:expa_expb}. The lowest values are in bold. In most of the sequences, it's shown that DSO is performing similarly using the photometric estimates obtaind from the two calibration methods, with an accuracy that is close to the full ground truth case. In addition to not being reliable on the CRF ground truth, the demonstrated accuracy of our method makes it the most applicable in real scenarios. To further prove this, and as a showcase of real applications, we modified Experiment \ref{itm:expb} by running the calibration method of \cite{engel2016photometrically}, this time without having access to the CRF ground truth. The estimates we acquired (shown in \cref{fig:their_est}) are then used as inputs for the DSO system. \cref{fig:trajectories_expb_expc} shows the trajectories obtained in one run of Experiments \ref{itm:expa}, \ref{itm:expb} and \ref{itm:expb} modified, on sequence 50. Notice how the quality of the trajectory degrades when less accurate photometric parameters are used.

\begin{table}[ht!]
\centering
\begin{adjustbox}{max width=\textwidth/2 - 11pt}
\begin{tabular}{ |c|c|c|c|c| } 
  \hline
  Seq. N. & Full GT & Our Calibration \ref{itm:expa} & Calibration \cite{bergmann2017online} \ref{itm:expb} & Without Calibration \ref{itm:expc} \\ 
  \hline
        8 & 0.3599 & 0.1638 & \textbf{0.1628} & 4.6209  \\ \hline
        16 & 0.5857 & \textbf{0.4539} & 0.4641 & 31.8692  \\ \hline
        24 & 0.2985 & \textbf{0.2864} & 0.3244 & 0.8757  \\ \hline
        25 & 0.9065 & 0.9136 & \textbf{0.9034} & 22.9724  \\ \hline
        28 & 1.7724 & 1.3649 & \textbf{1.1898} & 65.8894  \\ \hline
        29 & 0.3684 & \textbf{0.3602} & 0.4063 & 2.9593  \\ \hline
        30 & 0.7387 & 0.9056 & \textbf{0.6724} & 4.9369  \\ \hline
        31 & 0.6551 & \textbf{0.5371} & 0.7672 & 1.7746  \\ \hline
        32 & \textbf{0.2867} & 0.4412 & 0.3552 & 5.2303  \\ \hline
        35 & 0.7951 & \textbf{0.5120} & 0.9885 & 1.1007  \\ \hline
        36 & 3.6949 & \textbf{1.0530} & 2.1548 & 5.0325  \\ \hline
        37 & 0.4144 & \textbf{0.3420} & 0.4337 & 0.9906  \\ \hline
        38 & 0.5344 & \textbf{0.4860} & 1.0280 & 14.2914  \\ \hline
        39 & \textbf{1.5012} & 11.3534 & 4.5079 & 250.9258  \\ \hline
        40 & \textbf{1.0043} & 2.0538 & 2.6790 & 2.0116  \\ \hline
        41 & 0.3578 & 0.3811 & \textbf{0.3007} & 5.5642  \\ \hline
        42 & \textbf{0.4497} & 1.0604 & 1.2643 & 4.1518  \\ \hline
        43 & \textbf{0.3842} & 0.5191 & 0.4516 & 59.5276  \\ \hline
        44 & 0.8351 & 0.5768 & \textbf{0.5747} & 3.6844  \\ \hline
        45 & \textbf{1.2546} & 1.3091 & 1.2863 & 1.7220  \\ \hline
        46 & 0.6083 & 0.8033 & \textbf{0.5827} & 2.0079  \\ \hline
        48 & 0.8959 & \textbf{0.4531} & 0.6835 & 2.9357  \\ \hline
        49 & 0.9457 & \textbf{0.4976} & 0.7485 & 6.4594  \\ \hline
        50 & 0.8362 & \textbf{0.6118} & 0.6135 & 61.9891  \\ \hline
\end{tabular}
\end{adjustbox}
\caption{Trajectory alignment error $e_{align}$  calculated after running DSO on 24 TumMono sequences, using photometric parameters obtained from two calibration methods. Results are compared to both full ground truth (GT) and no photometric calibration cases.}
\label{table:expa_expb}
\end{table}

Furthermore, as shown in \cref{table:expa_expb}, disabling the photometric correction, or, in other words, forcing the brightness constancy assumption, led to the highest $e_{align}$ in 23 out of 24 times, which emphasizes the importance of photometric calibration for a direct VO system. Since our method operates online, processing frames as they arrive, it can be integrated into DSO or any other VO/SLAM system in future works.

\section{Ablation study}
\label{sec:ablation}

Although the estimation of vignette can be formulated as a simple least squares problem, different solvers will perform differently in the presence of noise and outliers. Here we examine the impact of different solving methods in enhancing the accuracy. We started by replacing the L2 norm with the L1 norm. The intuition behind this replacement is that the residuals would become subject to a linear loss rather than a quadratic one, decreasing the effect of outliers. Following the same concept, the M-estimator was tested and compared to L1 and L2 norm. The vignette RMSE values in the case of L2, L1, and M-estimator were 0.138, 0.0212, and 0.0184 respectively, demonstrating that using either of the L1 norm or the M-estimator instead of the L2 norm would decrease the RMSE by approximately 10 times.

The extraction of features is also a critical step since it directly affects the estimations. Several feature extractor methods were evaluated and compared based on the accuracy of the estimated CRF. SIFT, SURF, and ORB are the feature extractors that were tested. Replacing SIFT with the other two caused a performance drop, as the RMSE of the estimated CRF increased by $53\%$ (SURF) and $216\%$ (ORB).

The choice of the offset parameter, which specifies the number of frames to skip in the vignette estimation, is affected by camera motion. A slower camera motion necessitates a larger offset parameter to ensure sufficient radial movement, which we found to be greater than or equal to $20\%$ of the image diameter for better estimation. In our current implementation, we used a constant offset value of 30 frames on several TumMono sequences. To further test the sensitivity of this parameter, we first ran our vignette estimation on a video sequence from the challenging TumVI dataset \cite{schuberttum}. In this experiment, we noticed a comparable performance with the previous dataset, leading to a RMSE value of 0.018 when the offset is, again, at 30 frames. Then, we varied its value between 10 and 50 and analyzed the RMSE values. As a result, the variation showed a drop in the estimation accuracy when the offset was below 20 or above 32. In the case of the lower offset value (\eg, 20) a low number of matching points that do feature a high radial movement was found. Moreover, the higher the offset value, the lower the number of correct matches, leading to an unsolvable system when the camera has nearly left the scene.

In the CRF estimation module, the range we chose for the exposure ratio (\textit{0.92} and \textit{1.08}) to select keyframe pairs, depends on two criteria. First, as explained in \cref{subsec:Optimization}, the ratio should not be equal to 1. As we move further, increasing or decreasing, the system is theoretically solvable, until no such exposure change appears in the sequence. In the TumMono sequences we tested, \textit{0.92} and \textit{1.08} are good fits to ensure a sufficient number of keyframe pairs.

Our OSPC approach works whenever the exposure metadata is accessible. Some existing datasets do not provide the exposure times, which imposes a limitation on our method. However, to the best of our knowledge, retrieving the exposure values is possible when the camera is available, which makes our method applicable to most online applications. 

\section{Conclusion}
\label{sec:conc}
In this paper, we propose a novel method for online photometric calibration. Our approach relies on a sequential formulation that estimates the CRF, vignette, and exposure values consecutively. All three estimation problems are linear and convex, which accelerates the estimation process. We achieved high accuracy in all the estimates and showed how our photometric parameters enhance the results of a Visual Odometry System. Our key advantage is that we do not use ground truth information, as the previous approaches do. Instead, we rely on exposure values that can be acquired from a camera. 

\section*{Acknowledgment}
This work was supported by the Didymos Horizon Europe project, grant number 101092875–DIDYMOS-XR.

{\small
\bibliographystyle{ieee_fullname}
\bibliography{egbib}

\begin{thebibliography}{10}\itemsep=-1pt

\bibitem{bergmann2017online}
Paul Bergmann, Rui Wang, and Daniel Cremers.
\newblock Online photometric calibration of auto exposure video for realtime
  visual odometry and slam.
\newblock {\em IEEE Robotics and Automation Letters}, 3(2):627--634, 2017.

\bibitem{debevec2008recovering}
Paul~E Debevec and Jitendra Malik.
\newblock Recovering high dynamic range radiance maps from photographs.
\newblock In {\em ACM SIGGRAPH 2008 classes}, pages 1--10. 2008.

\bibitem{engel2017direct}
Jakob Engel, Vladlen Koltun, and Daniel Cremers.
\newblock Direct sparse odometry.
\newblock {\em IEEE transactions on pattern analysis and machine intelligence},
  40(3):611--625, 2017.

\bibitem{engel2016photometrically}
Jakob Engel, Vladyslav Usenko, and Daniel Cremers.
\newblock A photometrically calibrated benchmark for monocular visual odometry.
\newblock {\em arXiv preprint arXiv:1607.02555}, 2016.

\bibitem{1541349}
D.B. Goldman and Jiun-Hung Chen.
\newblock Vignette and exposure calibration and compensation.
\newblock In {\em Tenth IEEE International Conference on Computer Vision
  (ICCV'05) Volume 1}, volume~1, pages 899--906 Vol. 1, 2005.

\bibitem{1240119}
M.D. Grossberg and S.K. Nayar.
\newblock Determining the camera response from images: what is knowable?
\newblock {\em IEEE Transactions on Pattern Analysis and Machine Intelligence},
  25(11):1455--1467, 2003.

\bibitem{grossberg2003space}
Michael~D Grossberg and Shree~K Nayar.
\newblock What is the space of camera response functions?
\newblock In {\em 2003 IEEE Computer Society Conference on Computer Vision and
  Pattern Recognition, 2003. Proceedings.}, volume~2, pages II--602. IEEE,
  2003.

\bibitem{grundmann2013post}
Matthias Grundmann, Chris McClanahan, Sing~Bing Kang, and Irfan Essa.
\newblock Post-processing approach for radiometric self-calibration of video.
\newblock In {\em IEEE International Conference on Computational Photography
  (ICCP)}, pages 1--9. IEEE, 2013.

\bibitem{kanadeiterative}
Takeo Kanade.
\newblock An iterative image registration technique with an application to
  stereo vision (ijcai).

\bibitem{4587648}
Seon~Joo Kim, Jan-Michael Frahm, and Marc Pollefeys.
\newblock Radiometric calibration with illumination change for outdoor scene
  analysis.
\newblock In {\em 2008 IEEE Conference on Computer Vision and Pattern
  Recognition}, pages 1--8, 2008.

\bibitem{kim2010joint}
Seon~Joo Kim, David Gallup, Jan-Michael Frahm, and Marc Pollefeys.
\newblock Joint radiometric calibration and feature tracking system with an
  application to stereo.
\newblock {\em Computer Vision and Image Understanding}, 114(5):574--582, 2010.

\bibitem{kim2008robust}
Seon~Joo Kim and Marc Pollefeys.
\newblock Robust radiometric calibration and vignetting correction.
\newblock {\em IEEE transactions on pattern analysis and machine intelligence},
  30(4):562--576, 2008.

\bibitem{lin2004radiometric}
Stephen Lin, Jinwei Gu, Shuntaro Yamazaki, and Heung-Yeung Shum.
\newblock Radiometric calibration from a single image.
\newblock In {\em Proceedings of the 2004 IEEE Computer Society Conference on
  Computer Vision and Pattern Recognition, 2004. CVPR 2004.}, volume~2, pages
  II--II. IEEE, 2004.

\bibitem{lin2005determining}
Stephen Lin and Lei Zhang.
\newblock Determining the radiometric response function from a single grayscale
  image.
\newblock In {\em 2005 IEEE Computer Society Conference on Computer Vision and
  Pattern Recognition (CVPR'05)}, volume~2, pages 66--73. IEEE, 2005.

\bibitem{lowe2004distinctive}
David~G Lowe.
\newblock Distinctive image features from scale-invariant keypoints.
\newblock {\em International journal of computer vision}, 60(2):91--110, 2004.

\bibitem{mitsunaga1999radiometric}
Tomoo Mitsunaga and Shree~K Nayar.
\newblock Radiometric self calibration.
\newblock In {\em Proceedings. 1999 IEEE computer society conference on
  computer vision and pattern recognition (Cat. No PR00149)}, volume~1, pages
  374--380. IEEE, 1999.

\bibitem{mo2019ambiguity}
Zhipeng Mo, Boxin Shi, Sai-Kit Yeung, and Yasuyuki Matsushita.
\newblock Ambiguity-free radiometric calibration for internet photo
  collections.
\newblock {\em IEEE Transactions on Pattern Analysis and Machine Intelligence},
  42(7):1670--1684, 2019.

\bibitem{schuberttum}
David Schubert, Thore Goll, Nikolaus Demmel, Vladyslav Usenko, J{\"o}rg
  St{\"u}ckler, and Daniel Cremers.
\newblock The tum vi benchmark for evaluating visual-inertial odometry. in 2018
  ieee.
\newblock In {\em RSJ International Conference on Intelligent Robots and
  Systems (IROS)}, pages 1680--1687.

\bibitem{323794}
Jianbo Shi and Tomasi.
\newblock Good features to track.
\newblock In {\em 1994 Proceedings of IEEE Conference on Computer Vision and
  Pattern Recognition}, pages 593--600, 1994.

\bibitem{zheng2008single}
Yuanjie Zheng, Stephen Lin, Chandra Kambhamettu, Jingyi Yu, and Sing~Bing Kang.
\newblock Single-image vignetting correction.
\newblock {\em IEEE transactions on pattern analysis and machine intelligence},
  31(12):2243--2256, 2008.

\end{thebibliography}
}

\end{document}